\documentclass[lettersize,journal,compsoc]{IEEEtran}
\usepackage{amsmath,amsfonts}
\usepackage{algorithmic}
\usepackage{algorithm}
\usepackage{array}
\usepackage[caption=false,font=normalsize,labelfont=sf,textfont=sf]{subfig}
\usepackage{textcomp}
\usepackage{stfloats}
\usepackage{url}
\usepackage{verbatim}
\usepackage{graphicx}
\usepackage{cite}
\usepackage[pagebackref,breaklinks,colorlinks]{hyperref}
\usepackage[capitalize]{cleveref}
\crefname{section}{Sec.}{Secs.}
\Crefname{section}{Section}{Sections}
\Crefname{table}{Table}{Tables}
\crefname{table}{Tab.}{Tabs.}
\crefname{algorithm}{Alg.}{Algs.}
\usepackage{multirow}
\usepackage{booktabs}
\usepackage{colortbl}
\usepackage{xcolor}
\usepackage{pifont}

\begin{document}

\title{Exploring the Untouched Sweeps for Conflict-Aware 3D Segmentation Pretraining}

\author{Tianfang Sun, \and
        Zhizhong Zhang, \and
        Xin Tan, \and
        Yanyun Qu, \and
        Yuan Xie
\thanks{Tianfang Sun, Zhizhong Zhang, Xin Tan, Yuan Xie are with the School of Computer Science and Technology, East China Normal University, Shanghai 200060, China (e-mail: stf@stu.ecnu.edu.cn; zzzhang@cs.ecnu.edu.cn; xtan@cs.ecnu.edu.cn; xieyuan8589@foxmail.com).}
\thanks{Yanyun Qu is with the Department of Computer Science and Technology, Xiamen University, Fujian, China (e-mail: yyqu@xmu.edu.cn)}
}



\maketitle

\begin{abstract}
LiDAR-camera 3D representation pretraining has shown significant promise for 3D perception tasks and related applications. However, two issues widely exist in this framework: 1) Solely keyframes are used for training. For example, in nuScenes, a substantial quantity of unpaired LiDAR and camera frames remain unutilized, limiting the representation capabilities of the pretrained network. 2) The contrastive loss erroneously distances points and image regions with identical semantics but from different frames, disturbing the semantic consistency of the learned presentations. In this paper, we propose a novel Vision-Foundation-Model-driven sample exploring module to meticulously select LiDAR-Image pairs from unexplored frames, enriching the original training set. We utilized timestamps and the semantic priors from VFMs to identify well-synchronized training pairs and to discover samples with diverse content. Moreover, we design a cross- and intra-modal conflict-aware contrastive loss using the semantic mask labels of VFMs to avoid contrasting semantically similar points and image regions. Our method consistently outperforms existing state-of-the-art pretraining frameworks across three major public autonomous driving datasets: nuScenes, SemanticKITTI, and Waymo on 3D semantic segmentation by +3.0\%, +3.0\%, and +3.3\% in mIoU, respectively. Furthermore, our approach exhibits adaptable generalization to different 3D backbones and typical semantic masks generated by non-VFM models.
\end{abstract}


\section{Introduction}\label{sec:intro}
Accurate 3D semantic segmentation plays a crucial role in autonomous driving \cite{feng2020deep,rizzoli2022multimodal,hu2023planning}.
Self-supervision through LiDAR-camera contrastive learning has become a research hotpoint \cite{sautier2022image,mahmoud2023self,pang2023unsupervised,liu2023segment,liu2021learning,huang2021spatio,wu2023spatiotemporal,nunes2023temporal} recently. This paradigm facilitates providing robust initializations for various downstream tasks, enabling the training of more effective models with reduced fine-tuning effort.

A common practice is to transfer the knowledge from well-pretrained image backbones \cite{chen2020improved,caron2021emerging,caron2020unsupervised} to the 3D networks through cross-modal consistency.
A typical example, SLidR \cite{sautier2022image}, encourages the representation of 3D point regions, denoted as superpoints, to be consistent with the corresponding representation of 2D image regions, denoted as superpixels, through cross-modal contrastive loss. Seal \cite{liu2023segment} integrated the cross-modal contrastive loss, leveraging the superpixels generated by powerful 2D Vision Foundation Models (VFMs), such as SAM \cite{Kirillov_2023_ICCV}, with a temporal consistency loss based on the clusters created by the unsupervised clustering method HDBSCAN \cite{campello2013density}.

\begin{figure*}[]
  \centering
  \includegraphics[width=0.90\linewidth]{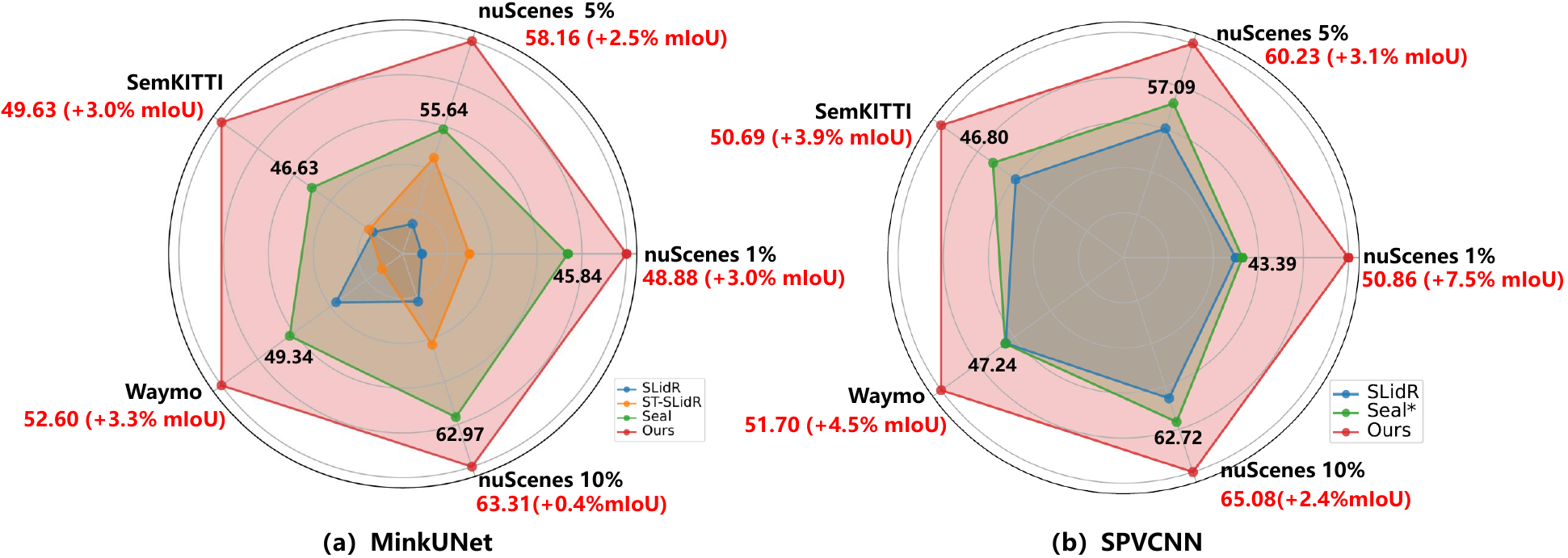}
  \caption{Comparison of few-shot finetuning performance on three public autonomous driving benchmarks \cite{caesar2020nuscenes,behley2019semantickitti,sun2020scalability} on two popular backbones \cite{choy20194d,tang2020searching}. The percentage after the dataset name denotes the percentage of annotation used for finetuning. Note that our methods are all pretrained on the nuScenes \cite{caesar2020nuscenes}. Our method significantly outperforms other recent cross-modal contrastive learning methods \cite{sautier2022image,mahmoud2023self,liu2023segment} on all three datasets under all the annotation settings.}
  \label{fig:intro}
\end{figure*}

Although promising results have been achieved, there remain two challenging problems: 1) \textit{Unexploited sweeps.} Current works \cite{sautier2022image,mahmoud2023self,liu2023segment,pang2023unsupervised} predominantly exploit keyframe set for training, leaving at least ten-fold quantity of unpaired LiDAR and camera frames, known as sweeps, untapped. Unlike keyframes, sweeps lack explicit 2D-3D matching relations and are therefore discarded. This underutilization of available data significantly constrains the representation ability of the pretrained network. 2) \textit{Cross-frame ``self-conflict'' problem.} Superpixels belonging to the same semantics but from different image frames are treated as negative samples. Although many attempts \cite{mahmoud2023self,liu2023segment} have been proposed, such as excluding the negative samples with features distances close to the anchor \cite{mahmoud2023self}, the cross-frame ``self-conflict'' issue remains unsolved and hampers the semantic consistency of the learned embeddings.

To address the first problem, we propose a VFM-driven sample exploring module. Initially, this module generates LiDAR-Image pairs by aligning each LiDAR frame with the images of the closest timestamps. This ensures that the LiDAR-Image pairs provide synchronized content. Subsequently, the temporally-aligned LiDAR-Image pairs with content most distinct from the nearby keyframes are selected, aided by semantic masks generated by VFMs \cite{oquab2023dinov2}.
This module facilitates the sampling of an additional training set from the previously unutilized sweeps by providing synchronized information and diverse content, thereby complementing the keyframe set. As a result, a total of 23,508 training samples have been added to the training set, doubling its size and consistently improving the downstream finetuning performance by +1.0 to +2.0 points in mIoU.

Secondly, we propose a conflict-aware method with cross/intra-modal contrastive loss. Our analysis reveals that current superpixel-driven contrastive learning methods are hindered by the cross-frame ``self-conflict'' problem. Remarkably, even provided with the ground truth as the superpixels, the improvement in finetuning performance remains modest using existing ``self-conflict'' contrastive loss. However, the segmentation masks generated by the VFMs, such as DINOv2 \cite{oquab2023dinov2}, typically assign the same mask ID to objects sharing similar semantics. Utilizing this prior, our cross-modal conflict-aware contrastive loss treats objects with the same mask ID as positive pairs, and others as negative pairs. This strategy effectively alleviates the cross-frame ``self-conflict'' problem. Similarly, we propose the intra-modal conflict-aware contrastive loss to enhance semantic consistency within the LiDAR modality, while simultaneously preventing the emergence of ``self-conflict''.

In conclusion, our contribution can be summarised as 1) We introduce a VFM-driven sample exploring module to utilize previously unused LiDAR and image frames, thereby doubling the training set without incurring extra costs. To the best of our knowledge, we are the first to utilize image frames from the sweeps in cross-modal pretraining. 2) We propose the cross-/intra-modal conflict-aware contrastive loss to mitigate the cross-frame ``self-conflict'' problem in both cross- and intra-modal domains. This approach fosters the learning of more semantically consistent embeddings, leading to enhanced finetuning performance across various downstream datasets. 3) Our approach demonstrates clear superiority over previous state-of-the-art methods in finetuning for the three most popular downstream autonomous driving datasets with diverse data configurations, as illustrated in \cref{fig:intro}.

\section{Related Work}\label{sec:rw}
\textbf{Vision Foundation Models.} Armed with massive training data \cite{Kirillov_2023_ICCV,radford2021learning} and advanced self-supervised learning techniques \cite{oquab2023dinov2,caron2021emerging}, some models exhibit promising zero-shot transfer capabilities across diverse downstream tasks. The segment anything model (SAM) \cite{Kirillov_2023_ICCV} demonstrates impressive generalization ability in various datasets for general-purpose image segmentation. Concurrently, models like OpenSeeD \cite{zhang2023simple}, SegGPT \cite{wang2023seggpt}, and DINOv2 \cite{oquab2023dinov2} show strong adaptability to a range of downstream tasks, such as image segmentation. In this work, we investigate methods for effectively transferring knowledge from these VFMs to 3D LiDAR semantic segmentation through conflict-aware contrastive learning.

\textbf{3D Presentation Learning.} Current methods can be mainly divided into two streams: mask-and-reconstruction-based \cite{zhang2023implicit,yang2023gd,tian2023geomae,boulch2023also,wu2023masked,xu2023mm,xu2023mv,chen2023pimae} and contrastive-based \cite{xie2020pointcontrast,sautier2022image,mahmoud2023self,liu2023segment,nunes2023temporal,wu2023spatiotemporal,peng2023openscene}. For the latter, some studies \cite{xie2020pointcontrast,zhang2021self,huang2021spatio,hou2021exploring} aim to train networks to learn features invariant to augmentations. However, these methods often struggle with limited data augmentation options for point clouds compared to images, posing challenges in terms of data diversity. Another stream of works focuses on transferring knowledge from pretrained image backbones by enforcing cross-modal consistency \cite{sautier2022image,mahmoud2023self,pang2023unsupervised,liu2023segment,peng2023openscene}. A common limitation of these approaches is the cross-frame ``self-conflict'' problem, as discussed earlier. We overcome this limitation with the proposed conflict-aware loss.

\textbf{Sweeps Utilization.} Most 3D contrastive learning methods primarily utilize the keyframe set for training. Several studies \cite{nunes2023temporal,wu2023spatiotemporal,huang2021spatio}, including Seal \cite{liu2023segment}, employ LiDAR sweeps alongside unsupervised clustering algorithms \cite{campello2013density,ester1996density}. Typically, these methods first cluster aggregated point clouds, then treat points from the same cluster but different timestamps as positive pairs for contrastive learning. TriCC \cite{pang2023unsupervised} enables the network to autonomously identify matching relationships among triplets formed from adjacent keyframes via the ``self-cycle''. However, these methods overlook the rich information available in images from the sweeps. Our method effectively utilizes both LiDAR and image frames from sweeps through the innovative VFM-driven sample exploring module.

\begin{figure*}[t]
  \centering
  \includegraphics[width=0.99\linewidth]{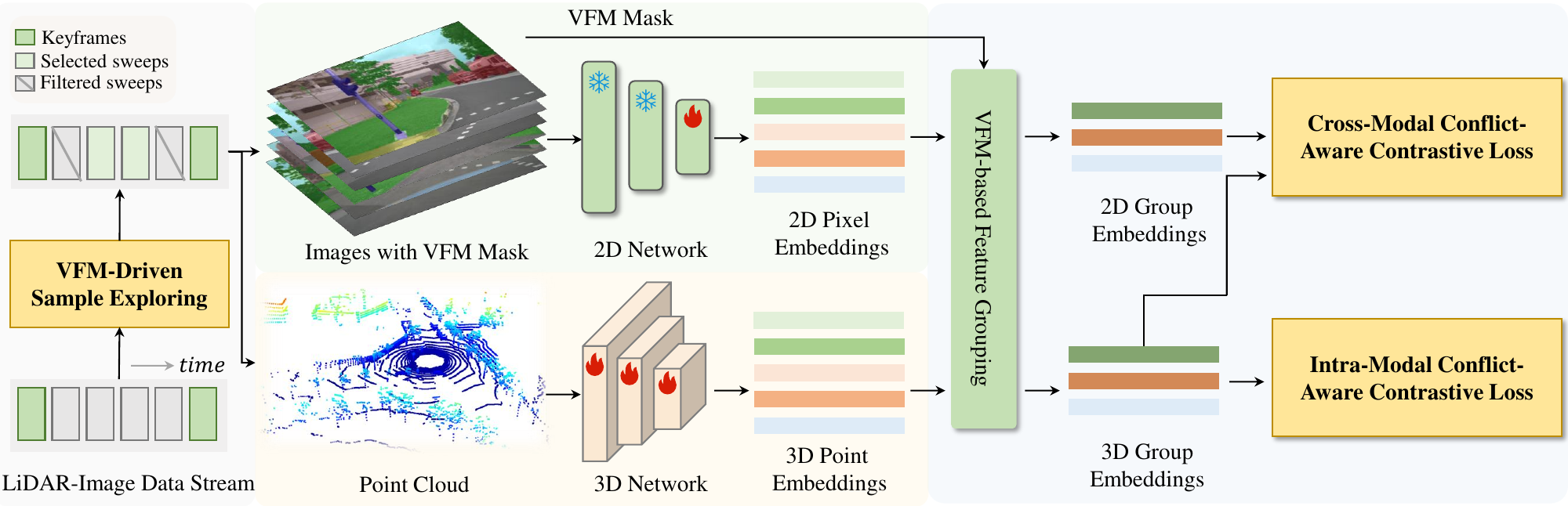}
  \caption{The overall frameworks. We sample both temporally-synchronized and distinct LiDAR-Image pairs from the untapped sweeps set with our VFM-driven sample exploring module (\cref{sec:vse}). The LiDAR-Image pairs are embedded into a unified feature space by the corresponding backbones and grouped by the VFM masks. The pretraining objective is conflict-aware contrastive learning (\cref{sec:ccl}), including the cross-modal conflict-aware contrastive loss and the intra-modal conflict-aware contrastive loss.}
  \label{fig:main}
\end{figure*}
\section{Method}\label{sec:method}
This paper considers the pretraining of a LiDAR network through cross-modal self-supervision from a pretrained image backbone (\textit{e.g.} a 2D vision foundation model). Subsequently, the pretrained LiDAR network serves as an effective initialization for various downstream datasets, enabling more satisfactory performance with reduced finetuning data.
Let \{${\boldsymbol{P}, \boldsymbol{I}}$\} represent a synchronized cross-modal input sample, where $\boldsymbol{P}\in \mathbb{R}^{N\times L}$ denotes the point cloud containing $N$ points with $L$-dimensional input (\textit{e.g.} intensity and elongation). Meanwhile,  $\boldsymbol{I}\in \mathbb{R}^{N_{cam}\times 3\times H\times W}$ denotes the images captured by $N_{cam}$ synchronized cameras, with $H$, $W$ denoting the height and width of the images, respectively. 

\textbf{Conflict-Unaware Contrastive Learning.} Prior works \cite{sautier2022image,mahmoud2023self} initially cluster visually similar regions of images into a set of superpixels, via the unsupervised SLIC \cite{achanta2012slic} algorithm. 
Owing to the available 2D-3D sensor calibration parameters, points in the point cloud can be mapped onto the corresponding image plane, thus transferring this supervision to 3D networks is readily achieved.
Therefore, the pretrained image feature at the superpixel level serves as a supervisory training signal for the LiDAR network through superpixel-driven contrastive learning. Essentially, the training objective focuses on improving the feature similarity between each superpixel and its corresponding point group, denoted as superpoint, while concurrently distancing other point groups within the feature space. It can be formulated as:
\begin{equation}\label{eq:nce}
    \mathcal{L}_{NCE}=-\frac{1}{M}\sum\limits_{i=0}^{M}\log \frac{e^{<\boldsymbol{q}_i, \boldsymbol{k}_i>/\tau}}{e^{<\boldsymbol{q}_i, \boldsymbol{k}_i>/\tau} + \sum_{j\neq i} {e^{<\boldsymbol{q}_i, \boldsymbol{k}_j>/\tau}}},
\end{equation}
where $\boldsymbol{q}_i\in \boldsymbol{Q} \in \mathbb{R}^{M\times D}$, $\boldsymbol{k}_i\in \boldsymbol{K} \in \mathbb{R}^{M\times D}$, $<\cdot, \cdot>$ and $\tau$ denotes to the superpixel feature, superpoint feature, scalar product and temperature hyper-parameter, respectively. $M$ denotes the number of superpixels. The superpixel and superpoint features are computed through average pooling.

However, this pipeline is facing two challenges: 1) Unexploited sweeps. $\boldsymbol{q}$ and $\boldsymbol{k}$ in \cref{eq:nce} are only collected from the keyframe set, leaving at least ten-fold quantity of samples unused. 2) Cross-frame ``self-conflict'' problem. $\boldsymbol{q}$ and $\boldsymbol{k}$ from the same semantics are erroneously accumulated for the denominator in \cref{eq:nce}. To address these problems, as illustrated in \cref{fig:main}, we select well-synchronized and distinct LiDAR-Image pairs from the unexploited sweeps using the VFM-driven sample exploring module. Afterwards, we propose a novel conflict-aware contrastive learning loss based on semantic masks generated by VFMs \cite{oquab2023dinov2}.

\begin{figure*}[t]
  \centering
  \includegraphics[width=0.99\linewidth]{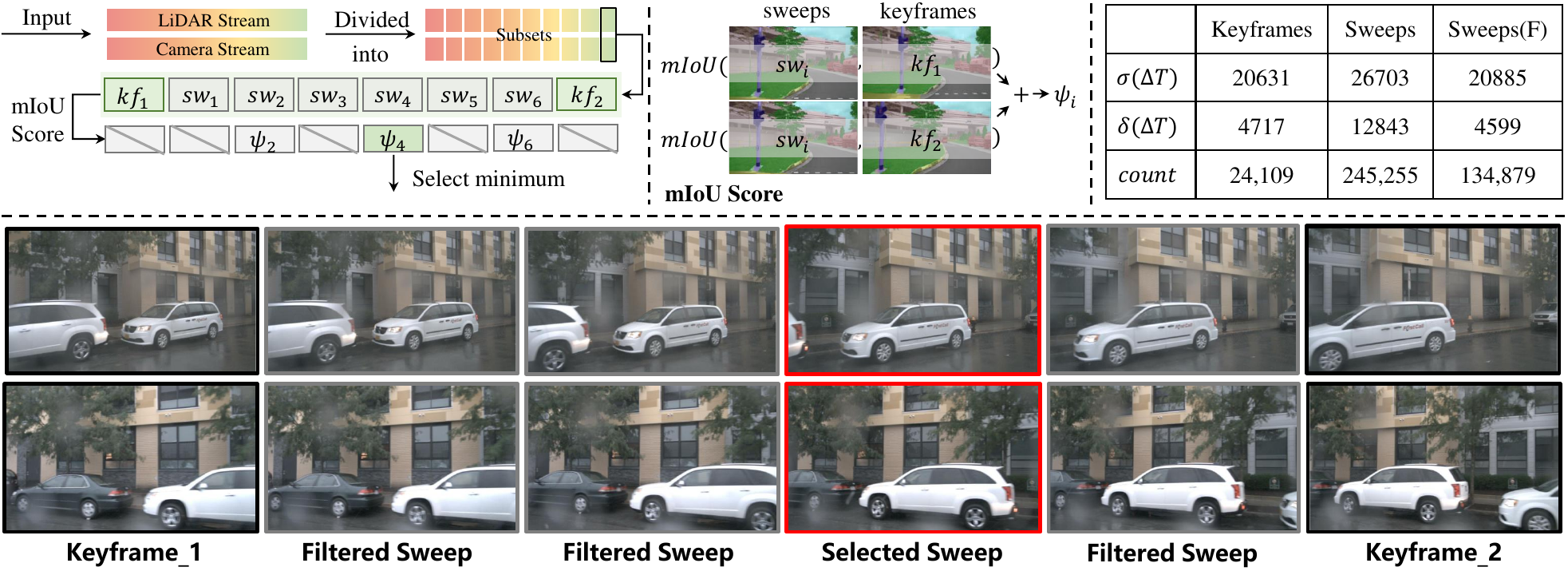}
  \caption{The VFM-driven sample exploring module. The top part of the figure is divided into three parts: Left: the overall pipeline of the proposed module. Middle: mIoU score calculation between any sweeps and keyframes. Right: The statistics of the sweeps selected by our module. The $\sigma(\Delta T)$ and $\delta(\Delta T)$ denote the mean and standard deviation of the timestamp difference. The $count$ represents the total number of samples in the corresponding set. The bottom part depicts two examples of the image of the selected sweeps (\textit{i.e.} images with red boundary).}
  \label{fig:vse}
\end{figure*}

\subsection{VFM-Driven Sample Exploring}\label{sec:vse}

There are two major obstacles hindering the effective use of the sweeps set in previous methods \cite{sautier2022image,mahmoud2023self,pang2023unsupervised,liu2023segment}: 1) The absence of matching relations. Unlike keyframes, whose matching relations between modalities are typically provided by datasets, such relations for sweeps are not readily available. 2) The potential similarity of content. The content within sweeps may appear too similar to the temporally adjacent keyframes to offer additional training benefits. To overcome these obstacles, we propose the VFM-driven sample exploring (VSE) module.

The comprehensive workflow of our VSE module is illustrated in \cref{fig:vse}. Taking the nuScenes \cite{caesar2020nuscenes} dataset for example, each LiDAR frame corresponds to six images captured by six cameras oriented in different directions around the vehicle. 
On average, there are 10 LiDAR frames and 60 images between two adjacent keyframes. These LiDAR frames and images form a ``subset'' as depicted in \cref{fig:vse}.
Our objective is to select LiDAR-Image pairs that are both temporally aligned and distinct in content (\textit{i.e.} select $sw_4$ from $sw_1$ to $sw_6$ in \cref{fig:vse}). This process contains the following two steps: \\

1) \textit{Build LiDAR-Image pairs}. Given a LiDAR frame, we compute its timestamp difference $\Delta T$ to every image. Subsequently, for each camera, the image exhibiting the smallest timestamp difference to the LiDAR frame is chosen as the corresponding image. Through this process, we construct LiDAR-Image pairs for each LiDAR frame. The average timestamp difference between the LiDAR frame and its corresponding images is denoted as $\sigma(sw)$ to be used in step (2). \\
2) \textit{Filter with predefined threshold}. As indicated in the ``Sweeps'' column in the right table in \cref{fig:vse}, the set built by step (1) exhibits a much larger mean and standard deviation in timestamp difference, compared to the keyframe set. This indicates some pairs are at a high risk of mismatch between the points and pixels.

To filter out these mismatched pairs, we compute the mean $\sigma$ and standard deviation $\delta$ of the timestamp difference of the keyframe set. The LiDAR-Image pairs with $\sigma(sw)$ less than $\sigma + \delta$ will be retained, as expressed by:
\begin{equation}
    \mathcal{S}_{sw} = \{ sw_h |\sigma(sw_h)=\frac{1}{N_{cam}}\sum\limits_{i=1}^{N_{cam}}\Delta T^{i} < \sigma + \delta \},
\end{equation}
with this approach, a subset $\mathcal{S}_{sw}$ of LiDAR-Image pairs is sampled from the sweeps (\textit{i.e.} $\{sw_2, sw_4, sw_6\}$ in \cref{fig:vse}), as depicted in the ``Sweeps(F)'' column in the right table in \cref{fig:vse}. 

Nevertheless, samples in $\mathcal{S}_{sw}$ continue to confront challenges related to contextual similarity with the keyframes.
To address this, we propose selecting the most distinct samples from the immediate preceding and succeeding keyframes (\textit{i.e.} $kf_1$ and $kf_2$) from $\mathcal{S}_{sw}$. A straightforward solution is to measure
the mIoU $\psi$ of the semantic masks generated by VFMs \cite{oquab2023dinov2} of images in $sw_h$ with the corresponding images in keyframes $kf_1$ and $kf_2$, as depicted in the middle of \cref{fig:vse}. The final similarity is calculated by the average mIoU of all of the images between it and the two keyframes. Finally, we select the samples with the lowest average mIoU (\textit{i.e.} $sw_4$ in \cref{fig:vse}) to form the additional training set.
With this approach, an additional 23,508 training samples are added to the training set in complement to the keyframes. We depict two examples of the image of the selected sweeps in the bottom \cref{fig:vse}. These training samples improve the variety of original training samples and therefore facilitate the finetuning performance on the downstream datasets. Empirical evidence suggests that merely selecting samples randomly from the sweeps without our VSE module, or simply extending the training epoch, can negatively impact the finetuning results.


\subsection{Conflict-Aware Contrastive Learning}\label{sec:ccl}

In this section, we present our conflict-aware contrastive learning.
We will commence with a proxy experiment designed to substantiate our assertion that the cross-frame ``self-conflict" problem exists in both cross-modal and intra-modal domains.

\begin{figure*}[h]
  \centering
  \includegraphics[width=0.99\linewidth]{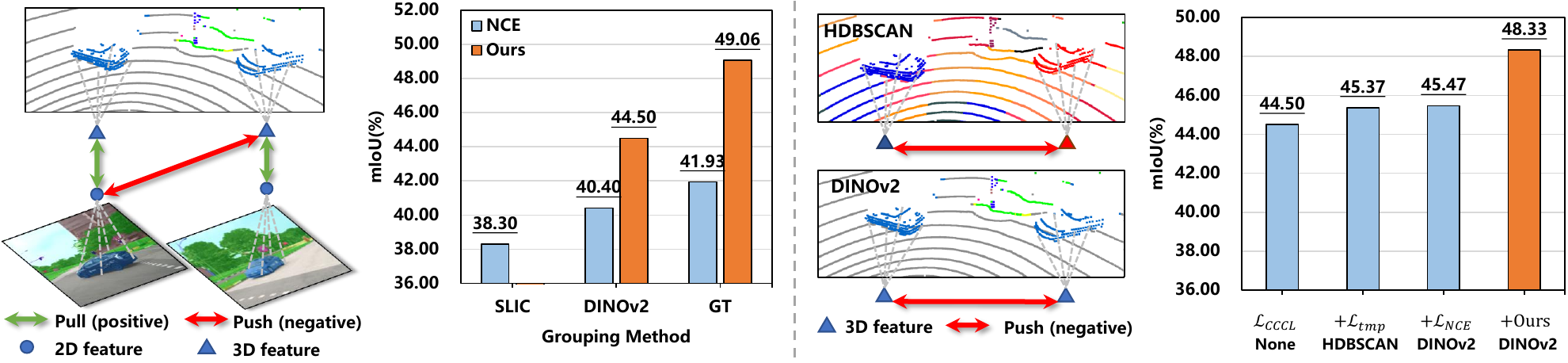}
  \caption{Proxy experiments. From left to right, the first and third figures depict the cross-frame ``self-conflict'' issue in cross- and intra-modal domains, respectively. The second and last figures report the proxy experiment results, where we report the 1\% finetune results on the nuScenes dataset after pretraining under different settings.}
  \label{fig:proxy}
\end{figure*}

\textbf{Proxy Experiment.} To demonstrate the ``self-conflict'' problem in cross-modal domain, we compare the finetuning performance of the weight pretrained by the cross-modal contrastive loss (\textit{i.e.} \cref{eq:nce}, denoted as NCE in \cref{fig:proxy}) used in the state-of-the-art method Seal \cite{liu2023segment} based on the superpixels generated by SLIC \cite{achanta2012slic}, DINOv2 \cite{oquab2023dinov2} and point-wise semantic ground truth label provided by the dataset. A critical observation is that despite a substantial improvement in superpixel quality from SLIC to the ground truth, the enhancement in finetuning performance is rather limited. However, the performance markedly improves when employing our conflict-aware method. This result leads us to conclude the cross-frame ``self-conflict'' problem, as shown in the left side of \cref{fig:proxy}, where objects of the same category from different frames are erroneously treated as negative samples, exists in the cross-modal domain.

Moreover, to illustrate the existence of the ``self-conflict'' problem within the intra-modal domain, we compare the finetuning performance of the models pretrained using the temporal-consistency loss from Seal \cite{liu2023segment} (\textit{i.e.} $\mathcal{L}_{tmp}$) based on the category-unaware clusters from HDBSCAN \cite{campello2013density} and the intra-modal version of the loss in \cref{eq:nce} relies on the corresponding point group from the category-aware mask from DINOv2. A key observation is that although the latter yields a more satisfactory clustering result, as depicted in the bottom left figure in \cref{fig:proxy}, it offers only marginal benefits to finetuning performance. A common drawback of both losses is their failure to address the ``self-conflict'' issue within the intra-modal domain. Therefore, it can be concluded that the ``self-conflict'' problem exists in both the cross- and intra-modal domains.

\textbf{Cross-Modal Conflict-Aware Contrastive Loss.} To begin with, we employ VFMs \cite{oquab2023dinov2} to produce the semantic mask set denoted as $\Phi_{\mathcal{S}}=\{(\boldsymbol{s}_i, l_i)|\,i=1,...,M\}$, where $\boldsymbol{s}\in\{0, 1\}^{H\times W}$ and $l$ denotes the semantic label. For DINOv2 \cite{oquab2023dinov2}, $l$ is in the label domain of ADE20K \cite{zhou2019semantic} as it is finetuned on the dataset. This is a common practice to obtain semantic masks. The corresponding point group $\Phi_{\mathcal{O}}=\{(\boldsymbol{o}_j, l_j)|\,j=1,...,M\}$ can be obtained based on the 2D-3D sensor calibration parameters. Let $F_{\theta_p}(\cdot): \mathbb{R}^{N\times L} \rightarrow \mathbb{R}^{N\times C_p}$ be a 3D encoder with trainable parameters $\theta_p$, that takes a LiDAR point cloud as input and outputs a $C$-dimensional per-point feature. Let $G_{\theta_i}(\cdot): \mathbb{R}^{H\times W\times 3} \rightarrow \mathbb{R}^{\frac{H}{s}\times\frac{W}{s}\times C_i}$ be an image encoder with parameters $\theta_i$, which is initialized from a set of 2D self-supervised pretrained parameters. 

To compute this cross-modal conflict-aware contrastive loss, we initially build trainable projection heads $H_{\omega_p}$ and $H_{\omega_i}$, which map the 3D point embeddings and 2D image embeddings into a shared $D$-dimensional embedding space. The point projection head $H_{\omega_p}: \mathbb{R}^{N\times C_p} \rightarrow \mathbb{R}^{N\times D}$ consist of a linear layer and an $l_2$-normalization layer. The image projection head $H_{\omega_i}: \mathbb{R}^{\frac{H}{s} \times \frac{W}{s} \times C_i} \rightarrow \mathbb{R}^{\frac{H}{s} \times \frac{W}{s} \times D}$ sequentially consists of a convolution layer with a kernel size of 1, an $l_2$-normalization layer and a upsample layer to resize the output feature map to match the dimensions of the input image.

Afterwards, we group the feature maps of image and point with the semantic masks $\Phi_{\mathcal{S}}$ and $\Phi_{\mathcal{O}}$ and obtain the corresponding grouped pixel embeddings $\boldsymbol{Q}'=\{(\boldsymbol{q}_i, l_i)|i=0,...,M\}$ and the grouped point embedding $\boldsymbol{K}'=\{(\boldsymbol{k}_j, l_j)|j=0,...,M\}$ by applying an average pooling function to each group of points and pixels, where $l_i$ and $l_j$ are the corresponding mask labels. The grouped features exhibit more robustness and avoid redundant calculations on trivial samples. Owning to the generalization ability of the VFMs, the masks with the same label usually belong to the same semantic categories. Therefore, inspired by \cite{khosla2020supervised}, we build the positive pair set for grouped points $\boldsymbol{k}_j$ as $\boldsymbol{P}(j)=\{\boldsymbol{q}_i\in \boldsymbol{Q}'| l_i = l_j\}$ with the guidance of the VFM labels. The formulation of the cross-modal conflict-aware contrastive loss is as follows:
\begin{equation}
    \mathcal{L}_{CCCL} = \sum\limits_{\boldsymbol{k}_j\in\boldsymbol{K}'}\frac{1}{|\boldsymbol{P}(j)|}\sum\limits_{\boldsymbol{q}_i\in \boldsymbol{P}(j)}\log\frac{e^{<\boldsymbol{k}_j, \boldsymbol{q}_i> / \tau}}{\sum\limits_{\boldsymbol{q}_z\in \boldsymbol{Q}'}{e^{<\boldsymbol{k}_j, \boldsymbol{q}_z> / \tau}}},
\end{equation}
where $|\boldsymbol{P}(j)|$ is the cardinality and $\tau$ is the temperature. 

\textbf{Intra-modal Conflict-Aware Contrastive Loss.} 
Intra-modal conflict-aware contrastive loss encourages the grouped points in $\boldsymbol{K}'$ of the same mask label $l_k$ to exhibit higher feature similarity, while simultaneously pushing apart grouped points with different mask labels. Therefore, we build the positive pair set for a grouped points $\boldsymbol{k}_j$ as $\boldsymbol{P}'(j)=\{\boldsymbol{k}_x \in \boldsymbol{K}'/\boldsymbol{k}_j | l_x=l_j\}$, where in avoid of trivial solution, we remove the grouped points itself from both the positive and negative set. The intra-modal conflict-aware contrastive loss is formulated as:
\begin{equation}
    \mathcal{L}_{ICCL} = \sum\limits_{\boldsymbol{k}_j\in\boldsymbol{K}'}\frac{1}{|\boldsymbol{P}'(j)|}\sum\limits_{\boldsymbol{k}_x\in \boldsymbol{P}'(j)}\log\frac{e^{<\boldsymbol{k}_j, \boldsymbol{k}_x> / \tau}}{\sum\limits_{\boldsymbol{k}_z\in \boldsymbol{K}'/\boldsymbol{k}_j}{e^{<\boldsymbol{k}_j, \boldsymbol{k}_z> / \tau}}},
\end{equation}
where $|\boldsymbol{P}'(j)|$ is the cardinality and $\tau$ is the temperature. 

\textbf{Discussion.} Our conflict-aware contrastive learning exhibits superiority over previous methods in three key aspects: 1) Our conflict-aware contrastive learning effectively addresses the ``self-conflict'' problem in both cross- and intra-modal domain, therefore, improving the semantic consistency of the learned feature. 2) The point group set is consistent in both $\mathcal{L}_{CCCL}$ and $\mathcal{L}_{ICCL}$, thereby avoiding potential optimization conflicts. 3) The finetuning performance of the models pretrained using our method significantly outperforms its counterpart, as illustrated in the orange bars in \cref{fig:proxy}.

\section{Experiments} \label{sec:exp}
\subsection{Experimental Setting}

\textbf{Datasets.} To extensively evaluate our proposed method, we pretrain all models on the nuScenes dataset and transfer our pretrained backbones to four different datasets: nuScenes \cite{caesar2020nuscenes}, SemanticKITTI \cite{behley2019semantickitti}, Waymo \cite{sun2020scalability}, and Synth4d \cite{saltori2022gipso}. \textbf{nuScene} is a large-scale public dataset for autonomous driving containing 700 training scenes. Each frame contains a 32-beam LiDAR point cloud provided with point-wise annotations, covering 16 semantic classes, and six RGB images captured by six cameras from different views of LiDAR. Following SLidR \cite{sautier2022image}, we further split the 700 scenes into 600 for pretraining and 100 for selecting the optimal hyper-parameters. \textbf{SemanticKITTI} is a large-scale dataset providing 43,000 scans with point-wise semantic annotations, Each scan contains a 64-beam LiDAR point cloud and the annotations cover 19 semantic categories. \textbf{Waymo} provides 23,691 frames for training and 5,976 frames for validating. Each frame contains a 64-beam LiDAR point cloud with point-wise annotations, which covers 22 semantic categories. \textbf{Synth4D} includes two subsets with point clouds captured by simulated Velodyne LiDAR sensors using the CARLA simulator. Following Seal \cite{liu2023segment}, we use the Synth4D-nuScenes subset in our experiments. It provides around 20,000 labeled point clouds, which cover 22 semantic categories. 

\textbf{Backbones.} We pretrain two 3D point cloud backbones with our method: Res16UNet implemented by Minkowski Engine \cite{choy20194d} and SPVCNN \cite{tang2020searching}. For Res16UNet, we keep all the settings the same as SLidR \cite{sautier2022image}. For SPVCNN, it takes input with Cartesian coordinates and the voxel size is $(0.1m, 0.1m, 0.1m)$ for $(x,y,z)-axis$. For the 2D backbone, we also keep all the same settings as SLidR by using the ResNet-50 \cite{he2016deep} with MoCov2 \cite{chen2020improved} as the pretrained parameters. We fix the parameters of ResNet and only train its projection heads.

\textbf{Implement Details.} During pretraining, all the backbones are pretrained on 4 RTX A6000 for 50 epochs using SGD with an initial learning rate of $1.6$, momentum of $0.9$, weight decay of $10^{-4}$. The batch size is 32 and a cosine annealing scheduler that decreases the learning rate from its initial value to 0 is adopted. The temperature $\tau$ is $0.07$ for both $\mathcal{L}_{CCCL}$ and $\mathcal{L}_{ICCL}$. For finetuning, we follow the exact augmentation and evaluation protocol as SLidR on nuScenes. As for SemanticKITTI and Waymo, we only change the batch size to 12 and kept the others the same as SLidR. The training objective is to minimize a combination of the cross-entropy loss and the Lovasz-Softmax loss \cite{berman2018lovasz}.

\begin{table*}[t]
\centering
\caption{Comparisons of different pretraining methods pretrained on \textit{nuScenes} \cite{caesar2020nuscenes} and finetuned on \textit{nuScenes} \cite{caesar2020nuscenes}, \textit{SemanticKITTI} \cite{behley2019semantickitti}, \textit{Waymo Open} \cite{sun2020scalability}, and \textit{Synth4D} \cite{saltori2022gipso}. \textbf{LP} denotes linear probing with frozen backbones. $\dagger$ denotes our re-implement. All mIoU scores are given in percentage (\%).}
\label{tab:nuscenes_semkitti}
\scalebox{0.99}{
\begin{tabular}{r|p{0.85cm}<{\centering}p{0.85cm}<{\centering}p{0.85cm}<{\centering}p{0.85cm}<{\centering}p{0.85cm}<{\centering}p{0.85cm}<{\centering}|p{1.00cm}<{\centering}|p{1.00cm}<{\centering}|p{1.00cm}<{\centering}}
\toprule
\multirow{2}{*}{\textbf{Method \& Year}} & \multicolumn{6}{c}{\textbf{nuScenes}} \vline & \textbf{KITTI} & \textbf{Waymo} & \textbf{Synth4D}
\\
& \textcolor{gray}{\textbf{LP}} & \textcolor{gray}{\textbf{1\%}} & \textcolor{gray}{\textbf{5\%}} & \textcolor{gray}{\textbf{10\%}} &  \textcolor{gray}{\textbf{25\%}} &  \textcolor{gray}{\textbf{Full}} & \textcolor{gray}{\textbf{1\%}} & \textcolor{gray}{\textbf{1\%}} & \textcolor{gray}{\textbf{1\%}} \\\midrule
\multicolumn{10}{l}{\textit{Res16UNet as the backbone}} \\
Random & $8.10$ & $30.30$ & $47.84$ & $56.15$ & $65.48$ & $74.66$ & $39.50$ & $39.41$ & $20.22$ \\
PointContrast~{\small\textcolor{gray}{[ECCV'20]}} \cite{xie2020pointcontrast} & $21.90$ & $32.50$ & - & - & - & - & $41.10$ & - & - \\
DepthContrast~{\small\textcolor{gray}{[ICCV'21]}} \cite{zhang2021self} & $22.10$ & $31.70$ & - & - & - & - & $41.50$ & - & - \\
PPKT~{\small\textcolor{gray}{[arXiv'21]}} \cite{liu2021learning}       & $35.90$ & $37.80$ & $53.74$ & $60.25$ & $67.14$ & $74.52$ & $44.00$ & $47.60$ & $61.10$ \\
SLidR~{\small\textcolor{gray}{[CVPR'22]}} \cite{sautier2022image}      & $38.80$ & $38.30$ & $52.49$ & $59.84$ & $66.91$ & $74.79$ & $44.60$ & $47.12$ & $63.10$ \\
ST-SLidR~{\small\textcolor{gray}{[CVPR'23]}} \cite{mahmoud2023self}    & $40.48$ & $40.75$ & $54.69$ & $60.75$ & $67.70$ & $75.14$ & $44.72$ & $44.93$ & -       \\
TriCC~{\small\textcolor{gray}{[CVPR'23]}} \cite{pang2023unsupervised}  & $38.00$ & $41.20$ & $54.10$ & $60.40$ & $67.60$ & $75.60$ & $45.90$ & -       & -       \\
Seal~{\small\textcolor{gray}{[NeurIPS'23]}} \cite{liu2023segment}      & $\mathbf{44.95}$ & $45.84$ & $55.64$ & $62.97$ & $68.41$ & $75.60$ & $46.63$ & $49.34$ & $64.50$ \\\midrule
\cellcolor{violet!7}\textbf{Ours}                                     & \cellcolor{violet!7}$40.00$ & \cellcolor{violet!7}$\mathbf{48.88}$ & \cellcolor{violet!7}$\mathbf{58.16}$ & \cellcolor{violet!7}$\mathbf{63.31}$ & \cellcolor{violet!7}$\mathbf{68.85}$ & \cellcolor{violet!7}$\mathbf{75.68}$ & \cellcolor{violet!7}$\mathbf{49.63}$ & \cellcolor{violet!7}$\mathbf{52.60}$ & \cellcolor{violet!7}$\mathbf{68.72}$ \\ \bottomrule
\multicolumn{10}{l}{\textit{SPVCNN as the backbone}} \\
SLidR$\dagger$~{\small\textcolor{gray}{[CVPR'22]}} \cite{sautier2022image}   & $44.50$ & $42.93$ & $55.77$ & $61.61$ & $69.84$ & $76.28$ & $45.10$ & $47.20$ & $63.29$ \\
Seal$\dagger$~{\small\textcolor{gray}{[NeurIPS'23]}} \cite{liu2023segment}  & $\mathbf{48.50}$ & $43.39$ & $57.09$ & $62.72$ & $70.40$ & $76.32$ & $46.80$ & $47.24$ & $62.07$ \\
\cellcolor{violet!7}\textbf{Ours}                                   & \cellcolor{violet!7}${48.10}$ & \cellcolor{violet!7}$\mathbf{50.86}$ & \cellcolor{violet!7}$\mathbf{60.23}$ & \cellcolor{violet!7}$\mathbf{65.08}$ & \cellcolor{violet!7}$\mathbf{70.78}$ & \cellcolor{violet!7}$\mathbf{76.38}$ & \cellcolor{violet!7}$\mathbf{50.69}$ & \cellcolor{violet!7}$\mathbf{51.70}$ & \cellcolor{violet!7}$\mathbf{66.60}$ \\
\bottomrule
\end{tabular}
}
\end{table*}

\subsection{Comparison to State-of-the-Arts}

We compare our method with random initialization and the most recent state-of-the-art methods on all four datasets to verify the superior performance and representation ability of our method. Two evaluation protocols are utilized for the pretrained models: 1) few-shot finetuning and 2) linear probing. For the former, we train the whole segmentation models with different proportions of available annotated training data to compare the annotation efficiency. For the linear probing, we initialize the parameters of the backbones with the pretrained weight and fix them, only training the linear segmentation head. The results are reported in \cref{tab:nuscenes_semkitti}. 

\textbf{On nuScenes.} For the few-shot finetuning, our method consistently outperforms Seal \cite{liu2023segment} under all the annotation settings using both two backbones. Specifically, when using Res16UNet, our method significantly outperforms Seal \cite{liu2023segment} by 3.0 and 2.5 at mIoU under 1\% and 5\% annotation settings. When using SPVCNN, the performance gain boosts to 7.5 and 3.1. These results showcase the importance of mining additional training samples from the sweeps and handling the ``self-conflict'' issue in both cross- and intra-modal, which results in a more representative and semantically consistent feature embedding. 
As for the linear probing results, we surpass SLidR \cite{sautier2022image} on both two backbones. Although non-ideal performance is found for the linear probing results, we believe the reason is that non-linear features are important in our method as \cite{pang2023unsupervised,he2022masked} shows. Also as illustrated in \cref{tab:sm_cmp}, when given the point-wise ground truth label as the semantic mask, the linear probing result of our method increases significantly to 54.57, surpassing its counterpart by 6.0 at mIoU, which indicates reducing the label noise in the VFM masks is beneficial for our method.

\textbf{On SemanticKITTI, Waymo, and Synth4D.} To showcase the representation ability of our method, we finetune our network on SemanticKITTI, Waymo, and Synth4D under 1\% annotation setting with the weight pretrained on nuScenes. The results are reported on the last three columns at \cref{tab:nuscenes_semkitti}. From the results, consistent improvements can be observed in all three datasets. Specifically, our method outperforms Seal \cite{liu2023segment} by 3.0, 3.3, and 4.2 at mIoU on SemanticKITTI, Waymo, and Synth4D, respectively. These results further reveal that the supplement of temporally aligned and diverse samples from sweeps and conflict-aware contrastive learning can improve the representation ability of the learned feature embedding.

\begin{table}[]
\centering
\caption{Ablation study of each component pretrained on \textit{nuScenes} \cite{caesar2020nuscenes} and finetuned on \textit{nuScenes} \cite{caesar2020nuscenes}, \textit{SemanticKITTI} \cite{behley2019semantickitti}, and \textit{Waymo Open} \cite{sun2020scalability}. $\mathcal{L}_{CCCL}$: Cross-modal conflict-aware contrastive loss. ${\mathcal{L}}_{ICCL}$: Intra-modal conflict-aware contrastive loss. \textbf{VSE}: VFM-driven sample exploring.}
\vspace{0.cm}
\label{tab:ablation}
\scalebox{0.8}{
\begin{tabular}{p{0.3cm}<{\centering}|p{0.72cm}<{\centering}p{0.72cm}<{\centering}p{0.72cm}<{\centering}|p{0.72cm}<{\centering}p{0.72cm}<{\centering}p{0.72cm}<{\centering}|p{1.0cm}<{\centering}|p{1.0cm}<{\centering}}
\toprule
\multirow{2}{*}{\textbf{\#}} & \multirow{2}{*}{$\mathcal{L}_{CCCL}$} & \multirow{2}{*}{$\mathcal{L}_{ICCL}$} & \multirow{2}{*}{\textbf{VSE}} & \multicolumn{3}{c}{\textbf{nuScenes}} \vline & \textbf{KITTI} & \textbf{Waymo} \\
& & & & \textcolor{gray}{\textbf{1\%}} & \textcolor{gray}{\textbf{5\%}} & \textcolor{gray}{\textbf{10\%}} & \textcolor{gray}{\textbf{1\%}} & \textcolor{gray}{\textbf{1\%}} \\\midrule
1) &  & & & $40.40$ & $55.00$ & $61.26$ & $46.28$ & $50.45$ \\ \midrule
2) & \checkmark & & & $44.50$ & $56.67$ & $61.84$ & $48.29$ & $51.17$ \\
3) & \checkmark & & \checkmark & $46.60$ & $57.86$ & $62.75$ & $47.84$ & $51.70$ \\
4) & \checkmark & \checkmark &  & $48.33$ & $58.05$ & $63.09$ &  $\mathbf{49.94}$ & $51.21$
\\\midrule
5) & \cellcolor{violet!7}\checkmark & \cellcolor{violet!7}\checkmark & \cellcolor{violet!7}\checkmark & \cellcolor{violet!7}$\mathbf{48.88}$ & \cellcolor{violet!7}$\mathbf{58.58}$ & \cellcolor{violet!7}$\mathbf{63.31}$ & \cellcolor{violet!7}$49.63$ & \cellcolor{violet!7}$\mathbf{52.60}$
\\
\bottomrule
\end{tabular}
}
\vspace{-0.4cm}
\end{table}

\subsection{Ablation Study}

\textbf{Component Analysis.} \cref{tab:ablation} presents an ablation study highlighting the contribution of each component. In the first row, we document the performance of SLidR \cite{sautier2022image} with the superpixels transitioned from SLIC \cite{achanta2012slic} to DINOv2 \cite{oquab2023dinov2} as the baseline. In Row\#2, we substitute the NCE loss \cite{sautier2022image} to our $\mathcal{L}_{CCCL}$. This change leads to an overall performance enhancement across all three datasets. Specifically, in the 1\% finetuning setting,  4.1, 2.0, and 0.7 mIoU improvements are observed in each dataset, respectively. These advancements are largely attributable to the effective mitigation of the cross-frame ``self-conflict'' issue in the cross-modal domain. In Row\#3, our VSE module is utilized, which yields overall improvements across these settings. These outcomes underscore the efficacy of our VSE module in extracting valuable training samples from the untapped sweeps set, without incurring additional costs. In Row\#4, our $\mathcal{L}_{ICCL}$ is incorporated. This addition, by enhancing the semantic consistency of intra-modal features, further elevates the finetuning performance. Specifically, there is a notable increase of 3.8 in mIoU in the 1\% annotation setting for the nuScenes dataset. Finally, in Row\#5, where all of our proposed modules are employed, the best results are achieved. This underscores that our modules are complementary to each other, collectively contributing to the enhanced performance. The non-ideal results on 1\% SemanticKITTI may be due to the coincidence of uniform sampling of training data.

\begin{table}
\centering
\caption{Semantic Mask Ablations. We compare the finetuning performance of our method with SLidR \cite{sautier2022image} based on various superpixel algorithms. The results are reported in mIoU on the validation set of nuScenes and SemanticKITTI.}
\label{tab:sm_cmp}
\resizebox{0.9\linewidth}{!}{
\begin{tabular}{@{}c|c|cc|c@{}}
\toprule
\multirow{2}{*}{Method} & \multirow{2}{*}{Superpixel} & \multicolumn{2}{c|}{nuScenes} & KITTI \\
                        &                             & LP            & 1\%           & 1\%   \\ \midrule
\multirow{4}{*}{SLidR\cite{sautier2022image}}  & SLIC \cite{achanta2012slic}  & 38.80         & 38.30         & 44.60 \\
                        & Oneformer \cite{jain2023oneformer}                  & 43.02         & 41.58         & 47.80 \\
                        & DINOv2 \cite{oquab2023dinov2}                       & 43.47         & 40.40         & 46.28 \\
                        & Ground Truth                                        & 48.60         & 41.93         & -     \\ \midrule
\multirow{3}{*}{Ours}   & Oneformer \cite{jain2023oneformer}                  & 37.90         & 44.17         & 48.10 \\
                        & DINOv2 \cite{oquab2023dinov2}                       & 40.35         & 48.33         & 49.94 \\
                        & Ground Truth                                        & 54.57         & 51.71         & -     \\ \bottomrule
\end{tabular}
}
\end{table}

\textbf{Semantic Mask Ablations.} To assess the generalizability of our method with semantic masks generated by non-VFM models, we conducted experiments using semantic masks from Oneformer \cite{jain2023oneformer}. Additionally, to determine the potential upper limit of this module's performance, we tested it using point-wise ground truth labels as semantic masks. Note that the VSE module is not used in this experiment. The results are reported in \cref{tab:sm_cmp}. From this data, two key observations emerge: 1) Our method remains effective even with the relatively noisy semantic masks produced by non-VFM models. 2) The finetuning performance of our method improves as the quality of the semantic masks increases. In comparison, the network pretrained by SLidR \cite{sautier2022image} does not exhibit such a trend, likely due to the persistent issue of cross-frame ``self-conflict''.


\begin{table}
\centering
\caption{Sweeps sampling comparisons. We compare the finetuning performance of various variants of our VSE module on sweeps selection strategies and pairing strategies. The results are reported in mIoU on the validation set of nuScenes.}\label{tab:vse}
\resizebox{0.9\linewidth}{!}{
\begin{tabular}{@{}c|ccc|ccc|c@{}}
\toprule
\multirow{2}{*}{} & \multicolumn{3}{c|}{Sweep selection}  & \multicolumn{3}{c|}{Pairing}   & \multirow{2}{*}{mIoU} \\
                  & MS         & Rand        & MD         & K2S        & S2K        & S2S        &                       \\ \midrule
Base              &            &             &            &            &            &            & 48.33                 \\
Var.1             & \checkmark &             &            &            &            & \checkmark & 47.72                 \\
Var.2             &            & \checkmark  &            &            &            & \checkmark & 47.52                 \\
Var.3             &            &             & \checkmark & \checkmark &            &            & 46.83                 \\
Var.4             &            &             & \checkmark &            & \checkmark &            & 47.94                 \\ \midrule
Ours              &            &             & \checkmark &            &            & \checkmark & 48.88                 \\ \bottomrule
\end{tabular}
}
\end{table}

\textbf{Sweeps Sampling Comparisons.} Comparative experiments were conducted between our proposed VSE module and several of its variants to validate its effectiveness. The results are presented in \cref{tab:vse}. Specifically, ``Base'' stands for our method without VSE module. ``MS'', ``Rand'', and ``MD'' signify selecting the most similar, random, and most distinct samples with the adjacent keyframes from the sweeps for pretraining. ``K2S'' denotes computing $\mathcal{L}_{CCCL}$ with the grouped pixel embedding from the keyframe and the grouped point embedding from the sweep selected by our VSE module. ``S2K'' is on the contrary. ``S2S'' stands for pairing the sweep points with sweep pixels.
From this data, we can draw two key conclusions: 1) When comparing our method with ``Var.1'' and ``Var.2'', our method shows superior performance, with improvements of 1.2 and 1.4, respectively. This underscores the importance of selecting samples with the most distinct content from the adjacent keyframes. 2) In comparisons with ``Var.3'' and ``Var.4'', our method outperforms them by 2.0 and 0.8, respectively, highlighting the significance of pairing the LiDAR and image frames in the sweeps with the most temporally-aligned counterparts. In summary, both steps in our VSE module are crucial for effectively constructing and utilizing LiDAR-Image pairs from the sweeps.

\begin{table}[]
\caption{Base Method Ablations. $^*$ stands for pretrained with DINOv2 superpixel. Seal$^*$ stands for our re-implementation due to unavailable official codes. The ``VSE'' stands for the using of our VSE module.}\label{tab:vse_abl}
\centering
\scalebox{0.99}{
\begin{tabular}{@{}c|c|ccc|c|c@{}}
\toprule
\multirow{2}{*}{Method}    & \multirow{2}{*}{\textbf{VSE}} & \multicolumn{3}{c|}{\textbf{nuScenes}} & \textbf{KITTI} & \textbf{Waymo} \\
                           &                      & \textcolor{gray}{\textbf{1\%}}      & \textcolor{gray}{\textbf{5\%}}      & \textcolor{gray}{\textbf{10\%}}    & \textcolor{gray}{\textbf{1\%}}   & \textcolor{gray}{\textbf{1\%}}   \\ \midrule
\multirow{2}{*}{SLidR$^*$} & \ding{53}            & 40.40    & 55.00    & 61.26   & 46.28 & 50.45 \\
                           & \checkmark           & \textbf{41.36}    & \textbf{55.14}    & \textbf{61.96}   & \textbf{46.47} & \textbf{51.04} \\ \midrule
\multirow{2}{*}{Seal$^*$}  & \ding{53}            & 41.17    & 55.41    & 61.83   & 47.06 & 50.51 \\
                           & \checkmark           & \textbf{42.06}    & \textbf{55.51}    & \textbf{62.64}   & \textbf{47.17} & \textbf{51.81} \\ \bottomrule
\end{tabular}
}
\end{table}

\textbf{Base Method Ablations.} To assess the generalizability of our VSE module with different base methods, we applied our VSE module to two prior works, \textit{i.e.} SLidR \cite{sautier2022image} and Seal \cite{liu2023segment}. For a fair comparison, these two methods were pretrained based on DINOv2 superpixels and we re-implemented Seal due to unavailable official codes. The results are depicted in \cref{tab:vse_abl}. From the table, the major conclusion can be drawn: our VSE module consistently improves the performance of SLidR and Seal on all these tested datasets, which validates the generalizability of our VSE module.

\subsection{Visualization}

\begin{figure*}[h]
  \centering
  \includegraphics[width=0.99\linewidth]{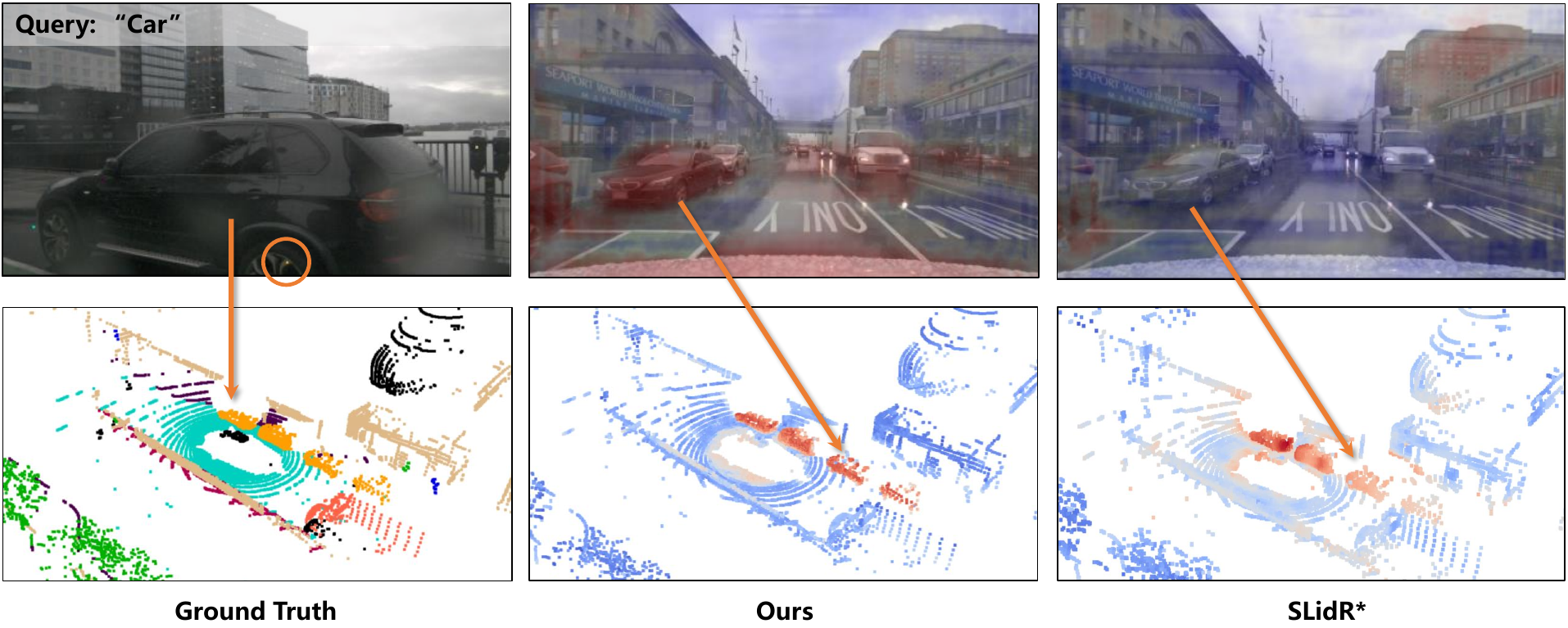}
  \caption{The intra- and cross-modal cosine similarity between a query point (emphasized by the orange circle) and the feature learned with our method and SLidR* \cite{sautier2022image}. The color goes from blue to red, denoting low and high similarity scores, respectively. The orange arrow points out the location of the objects in the images in the corresponding point cloud. In the top row, we calculate the cosine similarity between the 3D feature of the query point and the 2D feature of all pixels in the paired input sample. In the bottom row, we calculate the cosine similarity between the 3D feature of the query point and the entire point cloud.}
  \label{fig:vis_heatmap}
\end{figure*}

\textbf{Cosine Similarity.} As depicted in \cref{fig:vis_heatmap}, to demonstrate that our proposed method can eliminate the ``self-conflict'' issue and improve the semantic consistency of the pretrained features, we compute the intra- and cross-modal cosine feature similarity from a randomly selected 3D point to other points in the point cloud and all the pixels in the paired images. The results are shown in heat map style. For comparison, we also visualize the results from the model trained with SLidR \cite{sautier2022image} (\textit{i.e.} \cref{eq:nce}) based on the DINOv2 semantic masks. As observed, in both cross-modal and intra-modal contexts, our model exhibits higher feature similarity among the objects belonging to the same semantic categories, even when these objects originate from different image frames. However, the features learned by \cref{eq:nce} mainly focus on the objects within a single image, exhibiting less similarity with the objects from other images. This outcome indicates that our method facilitates the learning of more semantically-consistent features.

\begin{figure*}[h]
  \centering
  \includegraphics[width=0.99\linewidth]{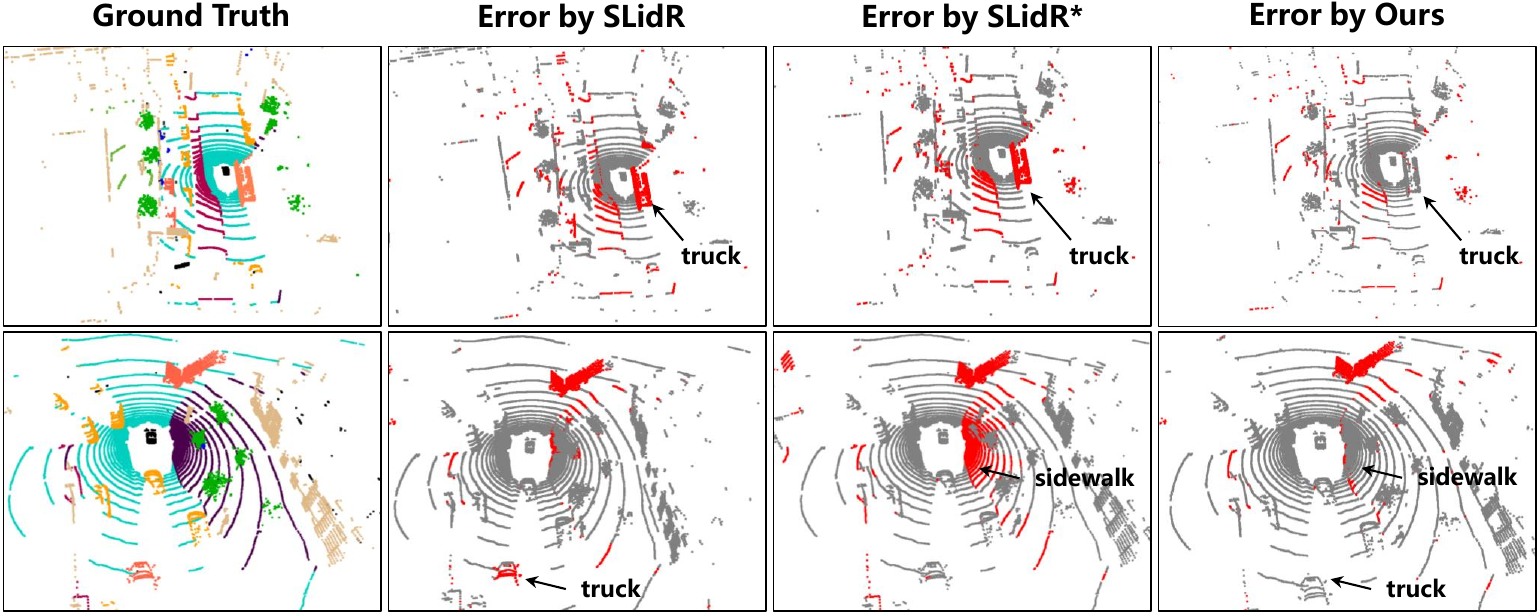}
  \caption{The qualitative results of different point cloud pretraining approaches pretrained on nuScenes and finetuned with 1\% labeled data. To highlight the differences, the correct/incorrect predictions are colored in gray/red, respectively. The results of SLidR* \cite{sautier2022image} are from our implementation based on the DINOv2 \cite{oquab2023dinov2} mask.}
  \label{fig:vis_pred_comp}
\end{figure*}

\textbf{Qualitative Analysis.} As shown in \cref{fig:vis_pred_comp}, we visualize the predictions of the network pretrained with the original SLidR \cite{sautier2022image}, the SLidR* based on DINOv2 superpixels, and our method on nuScenes \cite{caesar2020nuscenes} with 1\% annotated training samples for finetuning. Among all three tested methods, our method yields the best segmentation performance, which is credited to the more semantically consistent feature embeddings learned with our method.

\section{Conclusion}
In this paper, we investigate two problems aroused in LiDAR-Image contrastive learning (\textit{i.e.} unexploited sweeps and cross-frame ``self-conflict'' problem). To address these problems, we propose a VFM-driven sample exploring module and a cross-/intra-modal conflict-aware contrastive learning loss, which exploit previously untapped LiDAR-Image frames from the sweep set and address the ``self-conflict'' problem in both the cross- and intra-modal domain. As a result, our method achieves state-of-the-art performance on four datasets (\textit{i.e.} nuScenes, SemanticKITTI, Waymo, and Synth4D). We hope it provides new insights for the community of representation learning.

\bibliographystyle{splncs04}
\bibliography{main}

\begin{thebibliography}{10}
\providecommand{\url}[1]{\texttt{#1}}
\providecommand{\urlprefix}{URL }
\providecommand{\doi}[1]{https://doi.org/#1}

\bibitem{achanta2012slic}
Achanta, R., Shaji, A., Smith, K., Lucchi, A., Fua, P., S{\"u}sstrunk, S.: Slic superpixels compared to state-of-the-art superpixel methods. IEEE transactions on pattern analysis and machine intelligence  \textbf{34}(11),  2274--2282 (2012)

\bibitem{behley2019semantickitti}
Behley, J., Garbade, M., Milioto, A., Quenzel, J., Behnke, S., Stachniss, C., Gall, J.: Semantickitti: A dataset for semantic scene understanding of lidar sequences. In: Proceedings of the IEEE/CVF international conference on computer vision. pp. 9297--9307 (2019)

\bibitem{berman2018lovasz}
Berman, M., Triki, A.R., Blaschko, M.B.: The lov{\'a}sz-softmax loss: A tractable surrogate for the optimization of the intersection-over-union measure in neural networks. In: Proceedings of the IEEE conference on computer vision and pattern recognition. pp. 4413--4421 (2018)

\bibitem{boulch2023also}
Boulch, A., Sautier, C., Michele, B., Puy, G., Marlet, R.: Also: Automotive lidar self-supervision by occupancy estimation. In: Proceedings of the IEEE/CVF Conference on Computer Vision and Pattern Recognition. pp. 13455--13465 (2023)

\bibitem{caesar2020nuscenes}
Caesar, H., Bankiti, V., Lang, A.H., Vora, S., Liong, V.E., Xu, Q., Krishnan, A., Pan, Y., Baldan, G., Beijbom, O.: nuscenes: A multimodal dataset for autonomous driving. In: Proceedings of the IEEE/CVF conference on computer vision and pattern recognition. pp. 11621--11631 (2020)

\bibitem{campello2013density}
Campello, R.J., Moulavi, D., Sander, J.: Density-based clustering based on hierarchical density estimates. In: Pacific-Asia conference on knowledge discovery and data mining. pp. 160--172. Springer (2013)

\bibitem{caron2020unsupervised}
Caron, M., Misra, I., Mairal, J., Goyal, P., Bojanowski, P., Joulin, A.: Unsupervised learning of visual features by contrasting cluster assignments. Advances in neural information processing systems  \textbf{33},  9912--9924 (2020)

\bibitem{caron2021emerging}
Caron, M., Touvron, H., Misra, I., J{\'e}gou, H., Mairal, J., Bojanowski, P., Joulin, A.: Emerging properties in self-supervised vision transformers. In: Proceedings of the IEEE/CVF international conference on computer vision. pp. 9650--9660 (2021)

\bibitem{chen2023pimae}
Chen, A., Zhang, K., Zhang, R., Wang, Z., Lu, Y., Guo, Y., Zhang, S.: Pimae: Point cloud and image interactive masked autoencoders for 3d object detection. In: Proceedings of the IEEE/CVF Conference on Computer Vision and Pattern Recognition. pp. 5291--5301 (2023)

\bibitem{chen2020improved}
Chen, X., Fan, H., Girshick, R., He, K.: Improved baselines with momentum contrastive learning. arXiv preprint arXiv:2003.04297  (2020)

\bibitem{choy20194d}
Choy, C., Gwak, J., Savarese, S.: 4d spatio-temporal convnets: Minkowski convolutional neural networks. In: Proceedings of the IEEE/CVF conference on computer vision and pattern recognition. pp. 3075--3084 (2019)

\bibitem{ester1996density}
Ester, M., Kriegel, H.P., Sander, J., Xu, X., et~al.: A density-based algorithm for discovering clusters in large spatial databases with noise. In: kdd. vol.~96, pp. 226--231 (1996)

\bibitem{feng2020deep}
Feng, D., Haase-Sch{\"u}tz, C., Rosenbaum, L., Hertlein, H., Glaeser, C., Timm, F., Wiesbeck, W., Dietmayer, K.: Deep multi-modal object detection and semantic segmentation for autonomous driving: Datasets, methods, and challenges. IEEE Transactions on Intelligent Transportation Systems  \textbf{22}(3),  1341--1360 (2020)

\bibitem{he2022masked}
He, K., Chen, X., Xie, S., Li, Y., Doll{\'a}r, P., Girshick, R.: Masked autoencoders are scalable vision learners. In: Proceedings of the IEEE/CVF conference on computer vision and pattern recognition. pp. 16000--16009 (2022)

\bibitem{he2016deep}
He, K., Zhang, X., Ren, S., Sun, J.: Deep residual learning for image recognition. In: Proceedings of the IEEE conference on computer vision and pattern recognition. pp. 770--778 (2016)

\bibitem{hou2021exploring}
Hou, J., Graham, B., Nie{\ss}ner, M., Xie, S.: Exploring data-efficient 3d scene understanding with contrastive scene contexts. In: Proceedings of the IEEE/CVF Conference on Computer Vision and Pattern Recognition. pp. 15587--15597 (2021)

\bibitem{hu2023planning}
Hu, Y., Yang, J., Chen, L., Li, K., Sima, C., Zhu, X., Chai, S., Du, S., Lin, T., Wang, W., et~al.: Planning-oriented autonomous driving. In: Proceedings of the IEEE/CVF Conference on Computer Vision and Pattern Recognition. pp. 17853--17862 (2023)

\bibitem{huang2021spatio}
Huang, S., Xie, Y., Zhu, S.C., Zhu, Y.: Spatio-temporal self-supervised representation learning for 3d point clouds. In: Proceedings of the IEEE/CVF International Conference on Computer Vision. pp. 6535--6545 (2021)

\bibitem{jain2023oneformer}
Jain, J., Li, J., Chiu, M.T., Hassani, A., Orlov, N., Shi, H.: Oneformer: One transformer to rule universal image segmentation. In: Proceedings of the IEEE/CVF Conference on Computer Vision and Pattern Recognition. pp. 2989--2998 (2023)

\bibitem{khosla2020supervised}
Khosla, P., Teterwak, P., Wang, C., Sarna, A., Tian, Y., Isola, P., Maschinot, A., Liu, C., Krishnan, D.: Supervised contrastive learning. Advances in neural information processing systems  \textbf{33},  18661--18673 (2020)

\bibitem{Kirillov_2023_ICCV}
Kirillov, A., Mintun, E., Ravi, N., Mao, H., Rolland, C., Gustafson, L., Xiao, T., Whitehead, S., Berg, A.C., Lo, W.Y., Dollar, P., Girshick, R.: Segment anything. In: Proceedings of the IEEE/CVF International Conference on Computer Vision (ICCV). pp. 4015--4026 (October 2023)

\bibitem{liu2023segment}
Liu, Y., Kong, L., Cen, J., Chen, R., Zhang, W., Pan, L., Chen, K., Liu, Z.: Segment any point cloud sequences by distilling vision foundation models. arXiv preprint arXiv:2306.09347  (2023)

\bibitem{liu2021learning}
Liu, Y.C., Huang, Y.K., Chiang, H.Y., Su, H.T., Liu, Z.Y., Chen, C.T., Tseng, C.Y., Hsu, W.H.: Learning from 2d: Contrastive pixel-to-point knowledge transfer for 3d pretraining. arXiv preprint arXiv:2104.04687  (2021)

\bibitem{mahmoud2023self}
Mahmoud, A., Hu, J.S., Kuai, T., Harakeh, A., Paull, L., Waslander, S.L.: Self-supervised image-to-point distillation via semantically tolerant contrastive loss. In: Proceedings of the IEEE/CVF Conference on Computer Vision and Pattern Recognition. pp. 7102--7110 (2023)

\bibitem{nunes2023temporal}
Nunes, L., Wiesmann, L., Marcuzzi, R., Chen, X., Behley, J., Stachniss, C.: Temporal consistent 3d lidar representation learning for semantic perception in autonomous driving. In: Proceedings of the IEEE/CVF Conference on Computer Vision and Pattern Recognition. pp. 5217--5228 (2023)

\bibitem{oquab2023dinov2}
Oquab, M., Darcet, T., Moutakanni, T., Vo, H., Szafraniec, M., Khalidov, V., Fernandez, P., Haziza, D., Massa, F., El-Nouby, A., et~al.: Dinov2: Learning robust visual features without supervision. arXiv preprint arXiv:2304.07193  (2023)

\bibitem{pang2023unsupervised}
Pang, B., Xia, H., Lu, C.: Unsupervised 3d point cloud representation learning by triangle constrained contrast for autonomous driving. In: Proceedings of the IEEE/CVF Conference on Computer Vision and Pattern Recognition. pp. 5229--5239 (2023)

\bibitem{peng2023openscene}
Peng, S., Genova, K., Jiang, C., Tagliasacchi, A., Pollefeys, M., Funkhouser, T., et~al.: Openscene: 3d scene understanding with open vocabularies. In: Proceedings of the IEEE/CVF Conference on Computer Vision and Pattern Recognition. pp. 815--824 (2023)

\bibitem{radford2021learning}
Radford, A., Kim, J.W., Hallacy, C., Ramesh, A., Goh, G., Agarwal, S., Sastry, G., Askell, A., Mishkin, P., Clark, J., et~al.: Learning transferable visual models from natural language supervision. In: International conference on machine learning. pp. 8748--8763. PMLR (2021)

\bibitem{rizzoli2022multimodal}
Rizzoli, G., Barbato, F., Zanuttigh, P.: Multimodal semantic segmentation in autonomous driving: A review of current approaches and future perspectives. Technologies  \textbf{10}(4), ~90 (2022)

\bibitem{saltori2022gipso}
Saltori, C., Krivosheev, E., Lathuili{\'e}re, S., Sebe, N., Galasso, F., Fiameni, G., Ricci, E., Poiesi, F.: Gipso: Geometrically informed propagation for online adaptation in 3d lidar segmentation. In: European Conference on Computer Vision. pp. 567--585. Springer (2022)

\bibitem{sautier2022image}
Sautier, C., Puy, G., Gidaris, S., Boulch, A., Bursuc, A., Marlet, R.: Image-to-lidar self-supervised distillation for autonomous driving data. In: Proceedings of the IEEE/CVF Conference on Computer Vision and Pattern Recognition. pp. 9891--9901 (2022)

\bibitem{sun2020scalability}
Sun, P., Kretzschmar, H., Dotiwalla, X., Chouard, A., Patnaik, V., Tsui, P., Guo, J., Zhou, Y., Chai, Y., Caine, B., et~al.: Scalability in perception for autonomous driving: Waymo open dataset. In: Proceedings of the IEEE/CVF conference on computer vision and pattern recognition. pp. 2446--2454 (2020)

\bibitem{tang2020searching}
Tang, H., Liu, Z., Zhao, S., Lin, Y., Lin, J., Wang, H., Han, S.: Searching efficient 3d architectures with sparse point-voxel convolution. In: European conference on computer vision. pp. 685--702. Springer (2020)

\bibitem{tian2023geomae}
Tian, X., Ran, H., Wang, Y., Zhao, H.: Geomae: Masked geometric target prediction for self-supervised point cloud pre-training. In: Proceedings of the IEEE/CVF Conference on Computer Vision and Pattern Recognition. pp. 13570--13580 (2023)

\bibitem{wang2023seggpt}
Wang, X., Zhang, X., Cao, Y., Wang, W., Shen, C., Huang, T.: Seggpt: Segmenting everything in context. arXiv preprint arXiv:2304.03284  (2023)

\bibitem{wu2023masked}
Wu, X., Wen, X., Liu, X., Zhao, H.: Masked scene contrast: A scalable framework for unsupervised 3d representation learning. In: Proceedings of the IEEE/CVF Conference on Computer Vision and Pattern Recognition. pp. 9415--9424 (2023)

\bibitem{wu2023spatiotemporal}
Wu, Y., Zhang, T., Ke, W., S{\"u}sstrunk, S., Salzmann, M.: Spatiotemporal self-supervised learning for point clouds in the wild. In: Proceedings of the IEEE/CVF Conference on Computer Vision and Pattern Recognition. pp. 5251--5260 (2023)

\bibitem{xie2020pointcontrast}
Xie, S., Gu, J., Guo, D., Qi, C.R., Guibas, L., Litany, O.: Pointcontrast: Unsupervised pre-training for 3d point cloud understanding. In: Computer Vision--ECCV 2020: 16th European Conference, Glasgow, UK, August 23--28, 2020, Proceedings, Part III 16. pp. 574--591. Springer (2020)

\bibitem{xu2023mm}
Xu, M., Xu, M., He, T., Ouyang, W., Wang, Y., Han, X., Qiao, Y.: Mm-3dscene: 3d scene understanding by customizing masked modeling with informative-preserved reconstruction and self-distilled consistency. In: Proceedings of the IEEE/CVF Conference on Computer Vision and Pattern Recognition. pp. 4380--4390 (2023)

\bibitem{xu2023mv}
Xu, R., Wang, T., Zhang, W., Chen, R., Cao, J., Pang, J., Lin, D.: Mv-jar: Masked voxel jigsaw and reconstruction for lidar-based self-supervised pre-training. In: Proceedings of the IEEE/CVF Conference on Computer Vision and Pattern Recognition. pp. 13445--13454 (2023)

\bibitem{yang2023gd}
Yang, H., He, T., Liu, J., Chen, H., Wu, B., Lin, B., He, X., Ouyang, W.: Gd-mae: generative decoder for mae pre-training on lidar point clouds. In: Proceedings of the IEEE/CVF Conference on Computer Vision and Pattern Recognition. pp. 9403--9414 (2023)

\bibitem{zhang2023simple}
Zhang, H., Li, F., Zou, X., Liu, S., Li, C., Yang, J., Zhang, L.: A simple framework for open-vocabulary segmentation and detection. In: Proceedings of the IEEE/CVF International Conference on Computer Vision. pp. 1020--1031 (2023)

\bibitem{zhang2023implicit}
Zhang, Z., Bai, M., Li, E.: Implicit surface contrastive clustering for lidar point clouds. In: Proceedings of the IEEE/CVF Conference on Computer Vision and Pattern Recognition. pp. 21716--21725 (2023)

\bibitem{zhang2021self}
Zhang, Z., Girdhar, R., Joulin, A., Misra, I.: Self-supervised pretraining of 3d features on any point-cloud. In: Proceedings of the IEEE/CVF International Conference on Computer Vision. pp. 10252--10263 (2021)

\bibitem{zhou2019semantic}
Zhou, B., Zhao, H., Puig, X., Xiao, T., Fidler, S., Barriuso, A., Torralba, A.: Semantic understanding of scenes through the ade20k dataset. International Journal of Computer Vision  \textbf{127},  302--321 (2019)

\end{thebibliography}

\vfill

\end{document}